\documentclass{article}

\usepackage{microtype}
\usepackage{graphicx}
\usepackage{subfigure}
\usepackage{booktabs} 

\usepackage[dvipdfmx]{hyperref}

\usepackage[accepted]{icml2025}

\usepackage{amsmath}
\usepackage{amssymb}
\usepackage{mathtools}
\usepackage{amsthm}

\usepackage[capitalize,noabbrev]{cleveref}

\usepackage{bm}
\usepackage{amsmath,amssymb,mathrsfs}
\usepackage{graphicx}
\usepackage{comment}
\usepackage{multirow,array}
\usepackage{caption}
\usepackage{ulem}

\theoremstyle{plain}
\newtheorem{theorem}{Theorem}
\newtheorem{proposition}[theorem]{Proposition}

\theoremstyle{definition}

\theoremstyle{remark}

\icmltitlerunning{Kernel Method-based Kernel Intensity Estimators for Inhomogeneous Poisson Processes}

\begin{document}

\twocolumn[
\icmltitle{K$^2$IE: Kernel Method-based Kernel Intensity Estimators \\ for Inhomogeneous Poisson Processes}




\begin{icmlauthorlist}
\icmlauthor{Hideaki Kim}{ntt}
\icmlauthor{Tomoharu Iwata}{ntt}
\icmlauthor{Akinori Fujino}{ntt}
\end{icmlauthorlist}

\icmlaffiliation{ntt}{NTT Corporation, Japan}

\icmlcorrespondingauthor{Hideaki Kim}{hideaki.kin@ntt.com}

\icmlkeywords{point processes, Poisson processes, kernel methods, intensity estimators, least square losses}

\vskip 0.3in
]



\printAffiliationsAndNotice{}  

\begin{abstract}

Kernel method-based intensity estimators, formulated within reproducing kernel Hilbert spaces (RKHSs), and classical kernel intensity estimators (KIEs) have been among the most easy-to-implement and feasible methods for estimating the intensity functions of inhomogeneous Poisson processes. While both approaches share the term ``kernel", they are founded on distinct theoretical principles, each with its own strengths and limitations. In this paper, we propose a novel regularized kernel method for Poisson processes based on the least squares loss and show that the resulting intensity estimator involves a specialized variant of the representer theorem: it has the dual coefficient of unity and coincides with classical KIEs. This result provides new theoretical insights into the connection between classical KIEs and kernel method-based intensity estimators, while enabling us to develop an efficient KIE by leveraging advanced techniques from RKHS theory. We refer to the proposed model as the {\it kernel method-based kernel intensity estimator} (K$^2$IE). Through experiments on synthetic datasets, we show that K$^2$IE achieves comparable predictive performance while significantly surpassing the state-of-the-art kernel method-based estimator in computational efficiency.

\end{abstract}

\section{Introduction}
\label{intro}

Poisson processes have been the gold standard for modeling point patterns that occur randomly in multi-dimensional domains. They are characterized by an intensity function, that is, the instantaneous probability of events occurring at any point in the domain, which allows us to assess the risk of experiencing events at specified domains and forecast the timings/locations of future events. Poisson processes have a variety of applications in reliability engineering \cite{lai2006stochastic}, clinical research \cite{cox1972regression, clark2003survival, lanczky2021web}, seismology \cite{ogata1988statistical}, epidemiology \cite{gatrell1996spatial}, ecology \cite{heikkinen1999modeling}, and more.

Kernel intensity estimators (KIEs) are the simplest nonparametric approaches to estimating intensity functions \cite{ramlau1983smoothing, diggle1985kernel}, with advantages that include superior computational efficiency and theoretical tractability. They represent the underlying intensity function as a sum of smoothing kernels\footnote{Traditionally, the term ``kernel" is used to refer both to the weights assigned to data points in kernel density estimation and to the positive-definite kernels that define an RKHS. To avoid ambiguity, this paper employs two distinct terms ``smoothing" and ``RKHS" kernels, following \citet{flaxman17}.}\label{foot_kernel} evaluated at data points, where rescaled versions of density functions are usually adopted as smoothing kernels to correct the edge effects, that is, the estimation biases around the edges of observation domains. 

Recently, \citet{flaxman17} developed a feasible Reproducing Kernel Hilbert Space (RKHS) formulation for inhomogeneous Poisson processes. They showed that the representer theorem \cite{wahba90, scholkopf2001generalized} holds for a penalized maximum likelihood estimation under the constraint that the square root of the intensity function lies in an RKHS: the obtained square root of the intensity estimator is given by a linear combination of transformed RKHS kernels\footref{foot_kernel} evaluated at data points. The transformed RKHS kernels, often referred to as the {\it equivalent RKHS kernels}, naturally account for edge effects through likelihood functions, and the intensity estimator has been shown to outperform KIEs in scenarios where edge effects are prominent, such as in high-dimensional domains. Although the kernel method-based intensity estimator has a form similar to KIEs, it requires fitting the dual coefficient using a gradient descent method, making it less favorable than KIEs in terms of computational efficiency.

In this paper, we propose a penalized least squares loss formulation for estimating intensity functions under the constraint that the intensity function resides in an RKHS. The least squares loss is motivated by the empirical risk minimization principle \cite{geer2000empirical} and has demonstrated notable computational advantages in recent studies on Poisson processes \cite{hansen2015lasso, bacry2020sparse, cai2024latent}. Utilizing advanced variational analysis via path integral representation \cite{kim2021}, we show that a specialized variant of the representer theorem holds for the functional optimization problem: the resulting intensity estimator, which we call the {\it kernel method-based kernel intensity estimator} (K$^2$IE), has the unit dual coefficient and requires no optimization of dual coefficients given a kernel hyper-parameter, which is consistent with KIEs; furthermore, it employs the equivalent RKHS kernels appeared in Flaxman's model and can effectively address edge effects based on the RKHS theory. This result not only establishes a significant theoretical connection between classical KIEs and kernel method-based intensity estimators but also enables the development of a more scalable intensity estimator based on kernel methods.

In Section \ref{sec_back}, we outline related works of intensity estimation. In Section \ref{sec_prop}, we derive K$^2$IE via functional analysis with path integral representation of RKHS norm. In Section \ref{sec_experiment}, we compare K$^2$IE with conventional nonparametric intensity estimators on synthetic datasets, and confirm the effectiveness of the proposed method\footnote{Codes are available at: {\sf https://github.com/HidKim/K2IE} \label{foot_code}}. Finally, Section \ref{sec_conclusion} states our conclusions.

\section{Background}\label{sec_back}

Let a set of $N$ point events, $\mathcal{D} = \{ \bm{x}_n \}_{n=1}^N$, being observed in a $d$-dimensional compact space, $\mathcal{X} \subset \mathbb{R}^d$. We consider a learning problem of intensity function in the framework of inhomogeneous Poisson processes, where intensity function, $\lambda(\bm{x}): \mathcal{X} \rightarrow \mathbb{R}_+$, represents an instantaneous probability of events occurring at any point in $\mathcal{X}$:
\begin{equation}
\lambda(\bm{x}) = \lim_{|d\bm{x}|\rightarrow 0} \mathbb{E}[\mathcal{N}(d\bm{x})]~/~|d\bm{x}|,
\end{equation}
where $\mathcal{N}(d\bm{x})$ is the number of events occurring in $d\bm{x} \subset \mathcal{X}$, and $|\cdot|$ represents the measure of domain.

One of the most significant applications of Poisson processes lies in evaluating the risk of experiencing events (e.g., traffic accidents and disaster events) within specified regions. Given an intensity function $\lambda(\bm{x})$, the probability distribution of event counts (i.e., the Poisson distribution) over an arbitrary compact region $\mathcal{S} \subset \mathcal{X}$, denoted by $P_{\mathcal{S}}(\cdot)$, is calculated as follows:
\begin{equation}\label{eq_surv}
P_{\mathcal{S}}(n) = \frac{\Lambda^n e^{-\Lambda}}{n!}, \quad \Lambda = \int_{\mathcal{S}}\lambda (\bm{x}) d\bm{x},
\end{equation}
where $n \in \{ 0, 1, 2, \dots\}$. Using Equation (\ref{eq_surv}), we can perform binary classification to determine the occurrence of future events within $\mathcal{S}$. 

\subsection{Kernel Intensity Estimator}

Kernel smoothing is a classical approach to nonparametric intensity estimation \cite{diggle1985kernel}, expressed as:
\begin{equation}\label{eq_kie}
\hat{\lambda}(\bm{x}) = \sum_{n=1}^N g(\bm{x},\bm{x}_n) / \nu(\bm{x}), \ \ \nu(\bm{x}) = \int_{\mathcal{X}}g(\bm{x},\bm{s}) d\bm{s},
\end{equation}
where $g : \mathcal{X} \times \mathcal{X} \rightarrow \mathbb{R}_+$ represents a non-negative smoothing kernel\footref{foot_kernel}, and $\nu(\bm{x})$ is an edge-correction term\footnote{Intensity estimates near the edge of the domain $\mathcal{X}$ are biased downwards since no points are observed outside $\mathcal{X}$.}. This method, commonly referred to as the kernel intensity estimator (KIE), is closely related to the well-known kernel density estimator \cite{parzen1962estimation, davis2011remarks}, especially with bounded support \cite{jones1993simple}. KIEs offer several advantages, including theoretical tractability, superior computational efficiency, and ease of implementation. 

The smoothing kernel $g(\bm{x},\bm{x}')$ involves a bandwidth hyperparameter, which can be optimized using standard techniques such as cross-validation \cite{cronie2024cross} or Siverman's rule-of-thumb \cite{silverman2018density}. It is worth noting that widely-used cross-validation methods involving test log-likelihood functions require the integration of intensity functions over a test region $\mathcal{S} \subset \mathcal{X}$, where KIEs need to rely on time-consuming Monte Carlo integration because $g(\bm{x},\cdot)/\nu(\bm{x})$ in (\ref{eq_kie}) usually cannot be integrated in a closed form (e.g., Gaussian smoothing kernels). 

\subsection{Kernel Method-based Intensity Estimator}

Let $k : $ $\mathcal{X} \times \mathcal{X}$ $\rightarrow$ $\mathbb{R}$ denote a continuous positive semi-definite kernel. Then there exists a unique reproducing kernel Hilbert space (RKHS) $\mathcal{H}_k$ \citep{scholkopf2018learning, shawe2004kernel} associated with kernel $k(\cdot,\cdot)$.
\citet{flaxman17} modeled the intensity function as the square of a latent function in the RKHS, 
\begin{equation}
\lambda (\bm{x}) = f^2(\bm{x}), \quad f(\cdot) \in \mathcal{H}_k,
\end{equation}
and proposed a regularized minimization problem with log-likelihood loss functional as follows:
\begin{equation}
\min_{f \in \mathcal{H}_k} \biggl\{ - \sum_{n=1}^N \log(f^2 (\bm{x}_n)) + \int_{\mathcal{X}} f^2(\bm{x}) d\bm{x} + \frac{1}{\gamma} ||f||^2_{\mathcal{H}_k}\biggr\},
\end{equation}
where $||\cdot||^2_{\mathcal{H}_k}$ represents the squared Hilbert space norm, and $\gamma$ represents the regularization hyper-parameter. Through Mercer's theorem \cite{mercer1909xvi}, \citet{flaxman17} showed that the representer theorem \cite{wahba90, scholkopf2001generalized} does hold in an appropriately transformed RKHS, resulting in the following solution:
\begin{equation}\label{eq_kmie}
\hat{f}(\bm{x}) = \sum_{n=1}^N \alpha_n h(\bm{x},\bm{x}_n),
\end{equation}
where $\bm{\alpha} = (\alpha_1, \dots, \alpha_N)^{\top}$ is the dual coefficient, and $h : \mathcal{X} \times \mathcal{X} \rightarrow \mathbb{R}$ is a transformed RKHS kernel defined in terms of the Mercer expansion of $k(\bm{x},\bm{x}')$ as 
\begin{equation}\label{eq_equiv0}
\begin{split}
h(\bm{x}, \bm{x}') &= \sum_{m=1}^{\infty} \frac{\eta_m}{1/\gamma + \eta_m} e_m(\bm{x}) e_m(\bm{x}'), \\
\int_{\mathcal{X}} &k(\bm{x},\bm{s}) e_m(\bm{s}) d\bm{s} = \eta_m e_m(\bm{x}),
\end{split}
\end{equation}
where $\{e_m(\cdot)\}_{m=1}^{\infty}$ is the eigenfunctions of the integral operator $\int_{\mathcal{X}} \cdot k(\bm{x},\bm{s}) d\bm{s}$.
Recently, \citet{kim2022fastbayesian} rewrote the definition (\ref{eq_equiv0}) in terms of a Fredholm integral equation of the second kind \cite{polyanin98},
\begin{equation}\label{eq_equiv}
\frac{1}{\gamma} h(\bm{x},\bm{x}') + \int_{\mathcal{X}} k(\bm{x},\bm{s}) h(\bm{s},\bm{x}') d\bm{s} = k(\bm{x},\bm{x}'),
\end{equation}
which enables us to utilize the established approximation techniques for solving Fredholm integral equations \cite{polyanin98, atkinson2010personal}.
We will discuss how to solve Equation (\ref{eq_equiv}) in Section \ref{sec_equiv}. The transformed RKHS kernel is referred to as the {\it equivalent RKHS kernel} \cite{flaxman17, walder17, kim2022fastbayesian}.

The dual coefficient $\bm{\alpha}$ in the intensity estimator (\ref{eq_kmie}) solves the following dual optimization problem: 
\begin{equation}\label{eq_ppp}
\min_{\bm{\alpha} \in \mathbb{R}^N} \biggl\{ -\sum_{n=1}^N \log \sum_{n'=1}^N \alpha_{n'} h(\bm{x}_n,\bm{x}_{n'}) + \frac{1}{\gamma}\bm{\alpha}^{\top}\bm{H} \bm{\alpha} \biggr\},
\end{equation}
where $\bm{H} \coloneq [h(x_n, x_{n'})]_{nn'}$. The computational complexity of solving (\ref{eq_ppp}) is naively $\mathcal{O}(qN^2)$ for $q$ iterations of gradient descent methods, but reduces to $\mathcal{O}(qMN)$ when the equivalent RKHS kernel is given in degenerate form with rank $M$ ($<$ $N$) such that $ h(\bm{x},\bm{x}') = \sum_{m=1}^M \psi_m(\bm{x}) \psi_m(\bm{x}')$.

Flaxman's intensity estimator (\ref{eq_kmie}) differs from the classical KIE (\ref{eq_kie}) in that it does not require explicit edge correction. Instead, the equivalent RKHS kernel $h(\cdot, \cdot)$ naturally accounts for the effects of the finite observation domain $\mathcal{X} \subset \mathbb{R}^d$ through the second term on the left-hand side of Equation (\ref{eq_equiv}). \citet{flaxman17} demonstrated that the intensity estimator (\ref{eq_kmie}) outperforms the KIE in terms of predictive performance, especially in high-dimensional settings. However, it demands the model fitting (\ref{eq_ppp}) unlike KIEs, which makes KIEs more favorable regarding computational efficiency.

\subsection*{Other Related Works}

Gaussian Cox Processes (GCPs) provide a Bayesian alternative to kernel intensity estimators and kernel method-based models, where Gaussian processes are used to model latent intensity functions via positive-valued link functions \cite{moller98}. This framework enables principled interval estimation of intensity functions, along with hyperparameter inference in a fully Bayesian manner \cite{rathbun94, cunningham07, adams09, diggle13, gunter14, lloyd15, teng2017bayesian, donner18, john18, aglietti19}. Although GCPs are typically more computationally expensive than their non-Bayesian counterparts, extending kernel-based models into the Gaussian process framework can yield efficient Bayesian alternatives. For instance, \citet{walder17} proposed a Bayesian variant of Flaxman's model, known as the permanental process, and \citet{sellier2023sparse} further extended this approach within the generalized stationary kernel framework.

While neural network–based methods often trade computational efficiency for expressive power, \citet{tsuchida2024exact} recently introduced the squared neural family, a model that simultaneously achieves expressiveness and analytical tractability. Like Flaxman's model, it ensures non-negativity and closed-form expressiveness of the intensity function through the use of a squared link function--an elegant property that merits further attention.

\section{Method}\label{sec_prop}

\subsection{Kernel Method-based Kernel Intensity Estimator}\label{sec_k2ie}

In this paper, we introduce the least squares loss functional for Poisson processes \cite{hansen2015lasso} given by
\begin{equation}\label{eq_ls}
- 2\sum_{n=1}^N \lambda(\bm{x}_n) + \int_{\mathcal{X}} \lambda(\bm{x})^2 d\bm{x}.
\end{equation}
This loss functional comes from the empirical risk minimization principle \cite{geer2000empirical} and has demonstrated notable computational advantages in recent Poisson process literature \cite{hansen2015lasso, bacry2020sparse, cai2024latent}. For readers unfamiliar with the loss defined in (\ref{eq_ls}), we briefly explain the origin of the term {\it squares loss} in Appendix \ref{app_ls}. We model the intensity function as a latent function in RKHS $\mathcal{H}_k$, and consider the problem of minimization of the penalized least squares loss as follows:
\begin{equation}\label{eq_lsf}
\min_{\lambda \in \mathcal{H}_k} \biggl\{ - 2\sum_{n=1}^N \lambda(\bm{x}_n) + \int_{\mathcal{X}} \lambda(\bm{x})^2 d\bm{x} + \frac{1}{\gamma} ||\lambda||^2_{\mathcal{H}_k}\biggr\},
\end{equation}
where $||\cdot||^2_{\mathcal{H}_k}$ represents the squared Hilbert space norm, and $\gamma$ represents the regularization hyper-parameter. Through variational analysis, Theorem \ref{thm_k2ie} below shows that the resulting kernel method-based intensity estimator is consistent with the classical KIE (\ref{eq_kie}). While the main proof relies on the path integral representation \cite{kim2021}, for completeness, we also provide an alternative derivation based on Mercer's theorem in Appendix \ref{app_mercer}. To the best of our knowledge, this paper is the first to prove that the representer theorem holds for the penalized minimization of the least squares loss for the intensity estimation in RKHS. 

\begin{theorem}[] \label{thm_k2ie}
The solution of the functional optimization problem (\ref{eq_lsf}), denoted as $\hat{\lambda}(\cdot)$,  involves the representer theorem under a transformed RKHS kernel $h(\cdot,\cdot)$ defined by Equation (\ref{eq_equiv}), and its dual coefficient is equal to unity:
\begin{equation}\label{eq_k2ie}
\hat{\lambda}(\bm{x}) = \sum_{n=1}^N h(\bm{x},\bm{x}_n), \quad \bm{x} \in \mathcal{X}.
\end{equation}
\end{theorem}

\begin{proof}
Let $\mathcal{K}\cdot (\bm{x}) \!=\! \int_{\mathcal{X}} \cdot~k(\bm{x},\bm{s}) d\bm{s}$ be the integral operator with RKHS kernel $k(\cdot,\cdot)$, and $\mathcal{K}^* \! \cdot\!(\bm{x}) = \int_{\mathcal{X}} \cdot~k^*(\bm{x},\bm{s}) d\bm{s}$ be its inverse operator. Then, through the path integral representation of Gaussian processes \cite{kim2021}, the squared Hilbert space norm can be represented in a functional form,
 \begin{equation*}
 ||\lambda||^2_{\mathcal{H}_k}  = \iint_{\mathcal{X}\times\mathcal{X}} \!\!\!\! k^*(\bm{x},\bm{s}) \lambda(\bm{x}) \lambda(\bm{s}) d\bm{x} d\bm{s}. 
 \end{equation*}
 Using the representation, the objective functional in Equation (\ref{eq_lsf}) can be rewritten as follows:
\begin{equation*}
S (\lambda) = - 2\sum_{n=1}^N \lambda(\bm{x}_n) + \iint_{\mathcal{X}\times\mathcal{X}} \!\!\!\! q^*(\bm{x},\bm{s}) \lambda(\bm{x}) \lambda(\bm{s}) d\bm{x} d\bm{s},
\end{equation*}
where $q^*(\cdot,\cdot)$ is the weighted sum of $k^*(\cdot,\cdot)$ and the Dirac delta function $\delta(\cdot)$,
\begin{equation}\label{eq_equiv_inv}
q^*(\bm{x},\bm{s}) = \delta(\bm{x}-\bm{s})  + \frac{1}{\gamma}k^*(\bm{x},\bm{s}).
\end{equation}
The solution of  Equation~(\ref{eq_lsf}), $\hat{\lambda}(\bm{x})$, is obtained by solving the equation where the functional derivative of $S (\hat{\lambda})$ is equal to zero:
\begin{equation*}
\frac{\delta S}{\delta \hat{\lambda}} = -2 \sum_{n=1}^N \delta (\bm{x}-\bm{x}_n) + 2 \int_{\mathcal{X}} q^*(\bm{x},\bm{s}) \hat{\lambda}(\bm{s}) d\bm{s} = 0.
\end{equation*}
Let $\mathcal{Q}^*\!\cdot\!(\bm{x}) = \int_{\mathcal{X}} \cdot~q^*(\bm{x},\bm{s}) d\bm{s}$ be the integral operator associated with $q^*(\cdot,\cdot)$, and $\mathcal{Q}\!\cdot\!(\bm{x}) = \int_{\mathcal{X}} \cdot~q(\bm{x},\bm{s}) d\bm{s}$ be its inverse operator. Then applying operator $\mathcal{Q}$ to the equation, $\delta S/\delta \hat{\lambda} = 0$, leads to a representation of the form, 
\begin{equation*}
\hat{\lambda}(\bm{x}) = \sum_{n=1}^N q(\bm{x},\bm{x}_n),
\end{equation*}
where the relation, $(\mathcal{Q} \mathcal{Q}^*)\!\cdot\!(\bm{x}) =  \int_{\mathcal{X}} \cdot \delta(\bm{x}\!-\!\bm{s}) d\bm{s}$, was used. Furthermore, the following derivation shows that $q(\cdot,\cdot)$ is equal to the equivalent RKHS kernel $h(\cdot,\cdot)$ defined by (\ref{eq_equiv}): applying operator $\mathcal{Q}$ to Equation (\ref{eq_equiv_inv}) leads to the relation, 
\begin{equation*}
\delta(\bm{x}-\bm{x}') = q(\bm{x},\bm{x}') + \frac{1}{\gamma}  \int_{\mathcal{X}} k^*(\bm{x},\bm{s}) q(\bm{s},\bm{x}') d\bm{s},
\end{equation*}
and applying operator $\mathcal{K}$ to both sides of the relation yields:
\begin{equation*}
k(\bm{x},\bm{x}') = \int_{\mathcal{X}} k(\bm{x},\bm{s}) q(\bm{s},\bm{x}')d\bm{s} + \frac{1}{\gamma}  q(\bm{x},\bm{x}'),
\end{equation*}
which is identical to Equation (\ref{eq_equiv}).
\end{proof}

Theorem \ref{thm_k2ie} demonstrates, under the least squares loss functional, a strong connection between classical KIEs and modern kernel methods. From the perspective of KIE theory, Theorem \ref{thm_k2ie} implies that the equivalent RKHS kernels $h(\cdot,\cdot)$ are smoothing kernels constructed based on RKHS kernels. Hence, we call the proposed model (\ref{eq_k2ie}) the {\it kernel method-based kernel intensity estimator} (K$^2$IE). As \citet{flaxman17} discussed, the equivalent RKHS kernels implicitly incorporate edge effects in an effective manner. Therefore, our K$^2$IE is expected to combine the computational efficiency of KIEs with the effectiveness of Flaxman's kernel method-based estimator.

Similar to the conventional kernel method-based estimator (\ref{eq_kmie}), the support of K$^2$IE in Theorem \ref{thm_k2ie} lies within the observation domain $\mathcal{X}$, i.e., it concerns interpolation. However, by broadening the support of the RKHS kernel $k(\cdot, \cdot)$, the support in Theorem \ref{thm_k2ie} can be naturally extended: In other words, K$^2$IE defined by Equation~(\ref{eq_k2ie}) can be applied in its current form to extrapolation as well. A proof of this claim is provided in Appendix \ref{app_ext}.

Unlike conventional methods, K$^2$IE has the limitation of not guaranteeing the non-negativity of intensity functions. The equivalent RKHS kernels may generally take negative values, and since K$^2$IE is constructed as a linear combination of the equivalent RKHS kernels, it can yield negative values in certain regions, particularly in areas with no observed events. This issue is caused by the fact that K$^2$IE models intensity function by an RKHS function $f(\cdot) \in \mathcal{H}_{k}$, while conventional methods by $\sigma(f(\cdot))$ for a non-negative link function $\sigma(\cdot)$.  In practice, K$^2$IE does not have large negative values because the second term of the objective function (\ref{eq_lsf}) penalizes them. Thus we can deal with the issue by applying $\max(u,0)$ for intensity-related values $u$, such as $u = \lambda(\bm{x})$ and $u = \int_{\mathcal{S}} \lambda(\bm{x})d\bm{x}$ over a domain $\mathcal{S}$.

\subsection{Construction of Equivalent RKHS Kernel}\label{sec_equiv}

The primary task in K$^2$IE is to derive the equivalent RKHS kernel that satisfies the integral equation (\ref{eq_equiv}). The methodology varies depending on whether the observation domain $\mathcal{X}$ is infinite or finite, as elaborated in the subsequent sections. Here, we assume that RKHS kernels are shift-invariant, i.e., $k(\bm{x},\bm{x}')$ $=$ $k(\bm{x}-\bm{x}')$, which includes popular RKHS kernels such as Gaussian, Mat\'ern, and Laplace kernels.

\subsubsection{Infinite Observation Domain}

If the observation domain is infinite, i.e., $\mathcal{X} = \mathbb{R}^d$, the integral equation (\ref{eq_equiv}) can be solved by using the Fourier transform as follows:
\begin{equation}\label{eq_equiv_infinite}
h(\bm{x}-\bm{x}') = \mathcal{F}^{-1}\Biggl[ \frac{\tilde{k}(\bm{\omega})}{\gamma^{-1}+\tilde{k}(\bm{\omega})} \Biggr](\bm{x}-\bm{x}'),
\end{equation}
where $\mathcal{F}^{-1}[\cdot](\bm{x})$ denotes the inverse Fourier transform, and $\tilde{k}(\bm{\omega}\!\in\! \mathbb{R}^d)$ represents the Fourier transform of the shift-invariant RKHS kernel $k(\bm{x}\!-\!\bm{x}')$. Notably, the equivalent RKHS kernel $h(\cdot,\cdot)$ is also shift-invariant due to the symmetry of the integral equation (\ref{eq_equiv}). Approximation methods are required because the inverse Fourier transform in (\ref{eq_equiv_infinite}) generally cannot be expressed in closed form. One promising approach is the random feature map \cite{rahimi2007random}, where the equivalent RKHS kernel is approximated via Monte Carlo sampling from a probability distribution, $p(\cdot) \propto \tilde{k}(\cdot) / (\gamma^{-1}+\tilde{k}(\cdot))$, such that $h(\bm{x}\!-\!\bm{x}')$ $=$ $\mathbb{E}_{\bm{\omega} \sim p(\cdot)} [\exp(i\bm{w}^{\top}\bm{x})\exp(i\bm{w}^{\top}\bm{x}')]$. Another feasible approach is to apply the fast Fourier transform to (\ref{eq_equiv_infinite}).

When $\mathcal{X} = \mathbb{R}^d$, the edge-correction term in KIE (\ref{eq_kie}) vanishes, suggesting that the choice of the smoothing kernels $g(\bm{x},\bm{x}')$ in KIE (\ref{eq_kie}) is effectively equivalent to the selection of the equivalent RKHS kernels $h(\bm{x},\bm{x}')$ in K$^2$IE (\ref{eq_k2ie}). Through $h(\bm{x},\bm{x}')$, however, we could find smoothing kernels more robust to the squared error than popular ones such as Gaussian smoothing kernels. It is an interesting topic, but this paper focused on the case of a finite observation domain, where edge correction plays a crucial role.

\subsubsection{Finite Observation Domain}

\begin{figure*}[t]
\begin{center}
\centerline{\includegraphics[width=0.98\linewidth]{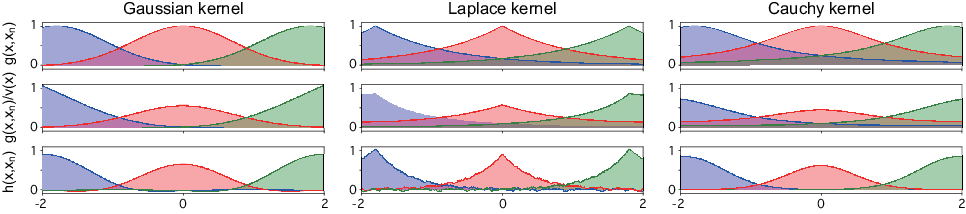}}
\caption{Examples of smoothing kernels $g(x,x_n)$, smoothing kernels with edge-correction $g(x,x_n)/\nu(x)$, and equivalent kernels $h(x,x_n)$ for data points $x_n \in \{ -1.8, 0, 1.8 \}$. Gaussian, Laplace, and Cauchy RKHS/smoothing kernels are $e^{-|x-x_n|^2}$, $e^{-|x-x_n|}$, and $\frac{1}{1+|x-x_n|^2}$, respectively. The regularization hyper-parameter, the number of random features, and the observation domain were set as $(\gamma, 2M, \mathcal{X}) = (2, 500, [-2,2])$.}
\label{fig_example}
\end{center}
\vskip -0.10in
\end{figure*}

Next, we consider a scenario where the observation domain is expressed as a union of a finite number of hyper-rectangular regions:
\begin{equation}\label{eq_region}
\mathcal{X} = \bigcup_{j=1}^J \mathcal{X}_j, \quad \mathcal{X}_j = \prod_{i=1}^d \bigl[ X^{\text{min}}_{ij}, X^{\text{max}}_{ij} \bigr],
\end{equation}
where $J$ denotes the number of hyper-rectangular regions, and $\mathcal{X}_j \cap \mathcal{X}_{j'\neq j} = \varnothing$. While prior studies typically assume a single hyper-rectangular region, the assumption (\ref{eq_region}) enables us to deal with more complicated observation domains, such as disjoint or irregularly shaped regions, often encountered in practical applications. 

The Fredholm integral equation (\ref{eq_equiv}) generally cannot be solved in closed form, and \citet{flaxman17} proposed using Nystr\"{o}m approximation \cite{williams2000using}, which approximates the integral term through numerical integration. While this approach has the advantage of being applicable to any RKHS kernel, it could potentially degrade the accuracy of the edge correction because the integral term is critical for the edge correction. To address the issue, we adopt the degenerate approach \cite{kim2022fastbayesian}, which approximates RKHS kernels using $2M$ random Fourier features \cite{rahimi2007random},
\begin{gather}
k(\bm{x}, \bm{x}') \simeq \sum_{m=1}^{2M} \phi_m(\bm{x}) \phi_m(\bm{x}') = \bm{\phi}(\bm{x})^{\top} \bm{\phi}(\bm{x}'), \\
\phi_m(\bm{x}) = M^{-1/2} \cos \bigl( \bm{\omega}_m^{\top} \bm{x} + \theta_m \bigr) \notag \\
\bm{\omega}_{m\leq M} \sim \tilde{k}(\bm{\omega}), \ \ \bm{\omega}_{m > M} = \bm{\omega}_{m-M}, \notag \\
\theta_{m\leq M} = 0,  \ \ \theta_{m > M} = -\pi/2, \notag
\end{gather}
and allows the integral term to be handled without any error as follows: 
\begin{equation}\label{eq_equiv_de}
\begin{split}
&h(\bm{x}, \bm{x}') = \bm{\phi}(\bm{x})^{\top} \bigl( \gamma^{-1} \bm{I}_{2M} + \bm{A} \bigr)^{-1}\bm{\phi}(\bm{x}'), \\
&\ \bm{A} = \sum_{j=1}^J \bm{A}^j, \quad \bm{A}^j =  \int_{\mathcal{X}_j} \bm{\phi}(\bm{x}) \bm{\phi}(\bm{x})^{\top} d\bm{x},
\end{split}
\end{equation}
where $\bm{I}_{2M}$ represents the identity matrix of size $2M$. Notably, $2M \!\times\! 2M$ matrix $\bm{A}$, which involves the edge-correction, can be computed in a closed form:
\begin{equation}
\begin{split}
(\bm{A}^j)_{mm'} &= \int_{\mathcal{X}_j} \phi_m(\bm{x}) \phi_{m'}(\bm{x}) d\bm{x} \\
&= \frac{1}{2M} \Bigl[ \zeta_j (\bm{\omega}_m+\bm{\omega}_{m'}, \theta_m+\theta_{m'}) \\
&\qquad \qquad + \zeta_j (\bm{\omega}_m-\bm{\omega}_{m'}, \theta_m-\theta_{m'}) \Bigr], \\
\zeta_j (\bm{\omega}, \theta) &= \cos \biggl[ \frac{1}{2} \sum_{i=1}^d \omega^i  \bigl( X_{ij}^{\text{max}}+X_{ij}^{\text{min}} \bigr) + \theta \biggr] \\
\cdot & \prod_{i=1}^d \bigl( X_{ij}^{\text{max}}-X_{ij}^{\text{min}} \bigr) \text{sinc} \biggl[ \frac{1}{2}\omega^i \bigl( X_{ij}^{\text{max}}-X_{ij}^{\text{min}} \bigr) \biggr],
\end{split}
\end{equation}
where $\text{sinc} (x)$ $=$ $\sin(x)/x$ is the unnormalized sinc function, and $\bm{\omega} \!=\! (\omega^1,\dots,\omega^d)^{\top}$. The relation (\ref{eq_equiv_de}) suggests that the equivalent kernel $h(\bm{x},\bm{x}')$ has degenerate form of rank $2M$, which is obtained through Cholesky decomposition as $h(\bm{x},\bm{x}') = (\bm{L}\bm{\phi}(\bm{x}))^{\top} (\bm{L}\bm{\phi}(\bm{x}'))$, where $\bm{L}^{\top} \bm{L} = (\gamma^{-1}\bm{I}_{2M} + \bm{A})^{-1}$.
To enhance the approximation accuracy of the random Fourier features, we employed the quasi-Monte Carlo feature maps \cite{yang2014quasi} in this paper.

The degenerate form of equivalent kernel (\ref{eq_equiv_de}) offers an additional advantage. For cross-validation with the least squares loss, K$^2$IE needs to evaluate the integral of the squared intensity function, $\int_{\mathcal{X}}(\sum_n h(\bm{x},\bm{x}_n))^2 d\bm{x}$, which requires $\mathcal{O}(N^2)$ computation naively or $\mathcal{O}(NZ)$ computation with $Z (\gg1) $ points Monte Carlo integration. But the integral of the squared intensity function can be obtained analytically with $\mathcal{O}(M^2 +MN)$ computation under (\ref{eq_equiv_de}) as follows:
\begin{equation}
\begin{split}
&\int_{\mathcal{X}} d\bm{x} \biggl[ \sum_{n} h(\bm{x},\bm{x}_n) \biggr]^2 = \bm{\xi}^{\top} \bm{A} \bm{\xi},  \\
& \bm{\xi} = \bigl( \gamma^{-1} \bm{I}_{2M} + \bm{A} \bigr)^{-1} \Bigl( \sum_n \bm{\phi}(\bm{x}_n) \Bigr).
\end{split}
\end{equation}
Therefore, regarding hyperparameter tuning, KIE and FIE require MC integration and solving a dual optimization problem for each cross-validation, respectively, whereas K$^2$IE requires neither, which is beneficial especially in multi-dimensional settings.

The comparison of K$^2$IE with KIE suggests that from the viewpoint of KIE, the primary distinction between them lies in how smoothing kernels with edge-correction are constructed: In KIE, the smoothing kernels with edge-correction are constructed by rescaling density functions with their integrals over the observation domain; In contrast, K$^2$IE constructs smoothing kernels as the solution to the integral equation (\ref{eq_equiv}). During model training, KIE benefits from more computational efficiency than K$^2$IE, which requires solving the integral equation. However, K$^2$IE offers computational advantages during inference as it can perform the intensity function integration needed in predictive tasks (e.g., see Equation (\ref{eq_surv})) analytically, while KIE relies on Monte Carlo integration. Furthermore, as demonstrated by \cite{flaxman17}, the smoothing kernel in K$^2$IE. i.e., the equivalent kernel is expected to achieve more effective edge correction, particularly in high-dimensional domain settings. 

Figure \ref{fig_example} illustrates examples of smoothing kernels with and without edge correction in KIE, as well as the equivalent RKHS kernels in K$^2$IE, showing that both the edge-corrected smoothing kernels and the equivalent RKHS kernels assign greater weight to data points near the boundary of the observation domain ($|x| \simeq 2$) compared to those at the center. Interestingly, K$^2$IE applies edge correction more conservatively through the equivalent RKHS kernels $h(x,x_n)$, that is, differentiates the weights between the center and the boundary less significantly compared to KIE with $g(x,x_n)/\nu(x)$.

\section{Experiments}\label{sec_experiment}

\begin{figure*}[t]
\begin{center}
\centerline{\includegraphics[width=0.98\linewidth]{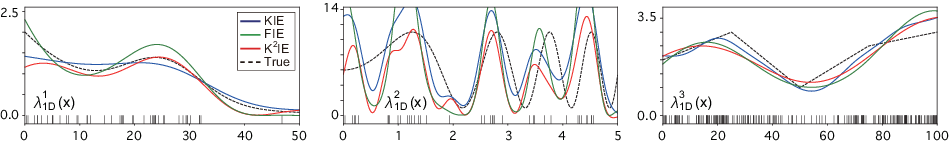}}
\caption{Examples of the estimated intensity functions on 1D synthetic data. The vertical lines represent the locations of observed events.}
\label{fig_1D}
\end{center}
\vskip -0.10in
\end{figure*}

\begin{table*}[]
\begin{center}
\caption{Results on 1D synthetic data across 100 trials with standard errors in brackets. $\tilde{N}$ denotes the average data size per trial. The performances not significantly ($p$ $>$ $10^{-2}$) different from the best one under the Mann-Whitney U test \cite{holm1979simple} are shown in bold for $L^2$ and $|L|$.}
\label{table_1d_synthetic}
\fontsize{9pt}{9.5pt}\selectfont{
\begin{tabular}{l>{\centering}p{--0.80cm}>{\centering}p{--0.80cm}>{\centering}p{--0.80cm}>{\centering}p{--0.80cm}>{\centering}p{--0.80cm}>{\centering}p{--0.80cm}>{\centering}p{--0.80cm}>{\centering}p{--0.80cm}>{\centering}p{--0.80cm}>{\centering}p{--0.80cm}>{\centering}p{--0.80cm}c@{}lc}
\toprule
\multirow{2}{*}{}&\multicolumn{4}{c}{$\lambda^1_{\text{1D}}(x): \tilde{N}\!=\!46$}&\multicolumn{4}{c}{$\lambda^2_{\text{1D}}(x): \tilde{N}\!=\!33$}&\multicolumn{4}{c}{$\lambda^3_{\text{1D}}(x): \tilde{N}\!=\!226$}\\ 
& $L^2 \!\downarrow$ & $|L| \!\downarrow$ & $\rho \!\uparrow$ & $cpu \!\downarrow$ & $L^2 \!\downarrow$ & $|L| \!\downarrow$ & $\rho \!\uparrow$ & $cpu \!\downarrow$ & $L^2 \!\downarrow$ & $|L| \!\downarrow$ & $\rho \!\uparrow$ & $cpu \!\downarrow$ \\ \midrule

\multirow{2}{*}{KIE} & \hspace{2.16pt}\bf{0.09} & \hspace{2.16pt}\bf{0.23} & -- & -- & \hspace{2.16pt}\bf{12.6} & \hspace{2.16pt}\bf{2.97} & -- & -- & \hspace{2.16pt}\bf{0.15} & \hspace{2.16pt}\bf{0.30} & -- & -- \\
 & (0.07) & (0.08) & -- & -- & (2.82) & (0.28) & -- & -- & (0.07) & (0.08) & -- & -- \\
\multirow{2}{*}{FIE} & \hspace{2.16pt}\bf{0.11} & \hspace{2.16pt}\bf{0.24} & \hspace{2.16pt}0.34 & \hspace{2.16pt}1.06 & \hspace{2.16pt}\bf{13.2} & \hspace{2.16pt}\bf{3.04} & \hspace{2.16pt}0.46 & \hspace{2.16pt}0.29 & \hspace{2.16pt}\bf{0.17} & \hspace{2.16pt}\bf{0.33} & \hspace{2.16pt}0.33 & 0.86 \\
 & (0.11) & (0.09) & -- & (0.17) & (3.41) & (0.28) & -- & (0.30) & (0.09) & (0.09) & -- & (0.39) \\
\multirow{2}{*}{K$^2$IE} & \hspace{2.16pt}0.12 & \hspace{2.16pt}0.26 & \hspace{2.16pt}0.26 & \hspace{2.16pt}0.01 & \hspace{2.16pt}\bf{13.9} & \hspace{2.16pt}\bf{3.09} & \hspace{2.16pt}0.48 & \hspace{2.16pt}0.01 & \hspace{2.16pt}0.18 & \hspace{2.16pt}0.34 & \hspace{2.16pt}0.31 & 0.01 \\
 & (0.08) & (0.08) & -- & (0.00) & (5.03) & (0.45) & -- & (0.00)  & (0.08) & (0.09) & -- & (0.00) \\

\toprule
\multirow{2}{*}{}&\multicolumn{4}{c}{$10\!\times\! \lambda^1_{\text{1D}}(x): \tilde{N}\!=\!466$}&\multicolumn{4}{c}{$10\!\times\!\lambda^2_{\text{1D}}(x): \tilde{N}\!=\!328$}&\multicolumn{4}{c}{$10\!\times\! \lambda^3_{\text{1D}}(x): \tilde{N}\!=\!2250$}\\ 
& $L^2 \!\downarrow$ & $|L| \!\downarrow$ & $\rho \!\uparrow$ & $cpu \!\downarrow$ & $L^2 \!\downarrow$ & $|L| \!\downarrow$ & $\rho \!\!\uparrow$ & $cpu \!\downarrow$ & $L^2 \!\downarrow$ & $|L| \!\downarrow$ & $\rho \!\uparrow$ & $cpu \!\downarrow$ \\ \midrule

\multirow{2}{*}{KIE} & \hspace{2.16pt}\bf{1.43} & \hspace{2.16pt}\bf{0.87} & -- &--  & \hspace{2.16pt}\bf{289} & \hspace{2.16pt}13.5 & -- & -- & \hspace{2.16pt}\bf{2.84} & \hspace{2.16pt}\bf{1.29} & -- & -- \\
 & (1.03) & (0.29) & -- & --  & (71.3) & (1.92) & -- & --  & (1.68) & (0.34) & -- & -- \\
\multirow{2}{*}{FIE} & \hspace{2.16pt}\bf{1.74} & \hspace{2.16pt}\bf{0.93} & \hspace{2.16pt}0.49 & \hspace{2.16pt}1.77 & \hspace{2.16pt}\bf{277} & \hspace{2.16pt}\bf{13.0} & \hspace{2.16pt}0.64 & \hspace{2.16pt}0.55 & \hspace{2.16pt}\bf{2.70} & \hspace{2.16pt}\bf{1.25} & \hspace{2.16pt}0.63 & 0.61 \\
 & (1.53) & (0.39) & -- & (0.13) & (80.6) & (2.09) & -- & (0.33) & (1.79) & (0.37) & -- & (0.13) \\
\multirow{2}{*}{K$^2$IE} & \hspace{2.16pt}\bf{1.67} & \hspace{2.16pt}\bf{0.92} & \hspace{2.16pt}0.49 & \hspace{2.16pt}0.01 & \hspace{2.16pt}\bf{266} & \hspace{2.16pt}\bf{12.7} & \hspace{2.16pt}0.77 & \hspace{2.16pt}0.01 & \hspace{2.16pt}\bf{3.24} & \hspace{2.16pt}\bf{1.34} & \hspace{2.16pt}0.47 & 0.01 \\
 & (0.71) & (0.36) & -- & (0.00) & (74.6) & (1.98) & -- & (0.00) & (2.08) & (0.41) & -- & (0.00) \\

\toprule
\end{tabular}
} 
\end{center}
\vskip -0.1in
\end{table*}

We evaluated the validity and the potential efficiency of our proposed K$^2$IE by comparing it with prior nonparametric approaches, including the kernel intensity estimator with edge correction (KIE) \cite{diggle1985kernel} and Flaxman's kernel method-based intensity estimator (FIE) \cite{flaxman17}, using synthetic datasets. For K$^2$IE and FIE, the number of random features $2M$ was fixed at 500 (see Appendix \ref{app_ablation} for an ablation study on the feature number $2M$).

For both the smoothing and RKHS kernels, we employed a multiplicative Gaussian function, $z(\bm{x},\bm{x}') =  e^{-|\bm{\beta}\circ(\bm{x}-\bm{x}')|^2}$, where $\bm{\beta} = (\beta_1,\dots, \beta_d)^{\top}$ is the inverse scale hyper-parameter, and $\circ$ denotes the Hadamard product. KIE optimized the hyper-parameter $\bm{\beta}$ through 5-fold cross-validation based on the negative log-likelihood function; FIE optimized the hyper-parameters, $(\bm{\beta}, \gamma)$, using the same cross-validation procedure as KIE; For K$^2$IE, the hyper-parameters, $(\bm{\beta}, \gamma)$, were optimized via 5-fold cross-validation with the least squares loss function (\ref{eq_ls}). For all models, the Monte Carlo cross-validation with $p$-thinning \cite{cronie2024cross} was adopted, where $p$ was fixed at $0.6$. A $10\!\times\!10$ logarithmic grid search was conducted for $\gamma \in [0.1, 100]$ and $\bm{\beta} \in [0.1, 100] \cdot \overline{\bm{\beta}}$, where $  \overline{\bm{\beta}} = ( \overline{\beta}_1, \dots, \overline{\beta}_d)^{\top}$ for $\overline{\beta}_i = 1/\bigl[\max_j \bigl(X_{ij}^{\text{max}}\bigr)-\min_j \bigl(X_{ij}^{\text{min}}\bigr)\bigr]$. For FIE, the gradient descent algorithm {\it Adam} \cite{kingma2014} was employed to solve the dual optimization problem (\ref{eq_ppp}).

Predictive performance was assessed using the integrated squared error ($L^2$) and the integrated absolute error ($|L|$) \cite{kowalczuk1998integrated}, defined as follows:
\begin{equation}
\begin{split}
L^2 &= \frac{1}{|\mathcal{X}|}\int_{\mathcal{X}} \bigl( \lambda^*(\bm{x}) - \hat{\lambda}(\bm{x}) \bigr)^2 d\bm{x}, \\
|L| &= \frac{1}{\bigl| \mathcal{X}|}\int_{\mathcal{X}} |\lambda^*(\bm{x}) - \hat{\lambda}(\bm{x}) \bigr| d\bm{x},
\end{split}
\end{equation}
where $\lambda^*(\bm{x})$ and $\hat{\lambda}(\bm{x})$ denote the true and estimated intensity functions, respectively. Following \cite{flaxman17}, the fraction of times that $L^2$ is smaller than KIE across the trials, denoted by $\rho$, was also reported, where $\rho$ was not defined for KIE. Efficiency was evaluated based on the CPU time (in seconds), $cpu$, required to execute the model fitting given the optimized hyper-parameters.

All models were implemented using TensorFlow-2.10\footref{foot_code} and executed on a MacBook Pro equipped with a 12-core CPU (Apple M2 Max), with the GPU disabled.

\subsection{1D Synthetic Data}

\begin{figure*}[t]
\begin{center}
\centerline{\includegraphics[width=0.95\linewidth]{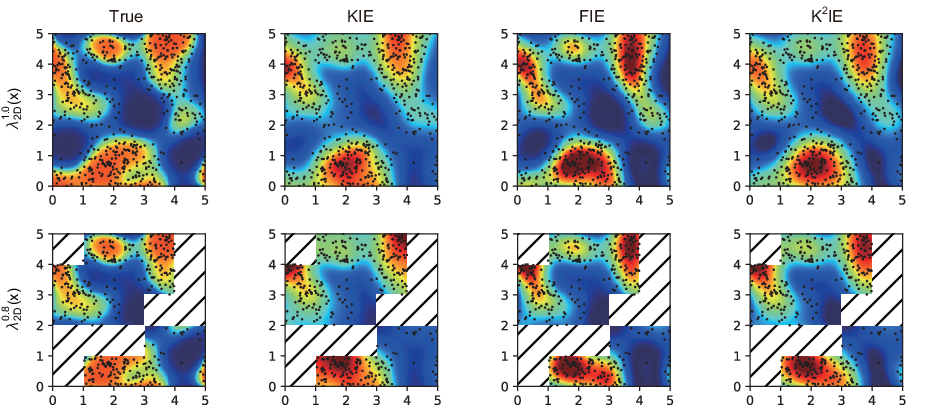}}
\caption{Examples of the estimated intensity functions on 2D synthetic data $\lambda_{\text{2D}}^{1.0}$ and $\lambda_{\text{2D}}^{0.8}$. The black dots represent the locations of observed events, and the unobserved regions are indicated by hatched lines.}
\label{fig_2D}
\end{center}
\end{figure*}

\begin{table*}[]
\begin{center}
\caption{Results on 2D synthetic data across 100 trials with standard errors in brackets. $\tilde{N}$ denotes the average data size per trial. The notation follows Table \ref{table_1d_synthetic}.}
\label{table_2d_synthetic}
\fontsize{9pt}{9.5pt}\selectfont{
\begin{tabular}{l>{\centering}p{--0.80cm}>{\centering}p{--0.80cm}>{\centering}p{--0.80cm}>{\centering}p{--0.80cm}>{\centering}p{--0.80cm}>{\centering}p{--0.80cm}>{\centering}p{--0.80cm}>{\centering}p{--0.80cm}>{\centering}p{--0.80cm}>{\centering}p{--0.80cm}>{\centering}p{--0.80cm}c@{}lc}
\toprule
\multirow{2}{*}{}&\multicolumn{4}{c}{$\lambda_{\text{2D}}^{1.0}(x): \tilde{N}\!=\!543$}&\multicolumn{4}{c}{$\lambda_{\text{2D}}^{0.9}(x): \tilde{N}\!=\!483$}&\multicolumn{4}{c}{$\lambda_{\text{2D}}^{0.8}(x): \tilde{N}\!=\!428$}\\ 
& $L^2 \!\downarrow$ & $|L| \!\downarrow$ & $\rho \!\uparrow$ & $cpu \!\downarrow$ & $L^2 \!\downarrow$ & $|L| \!\downarrow$ & $\rho \!\uparrow$ & $cpu \!\downarrow$ & $L^2 \!\downarrow$ & $|L| \!\downarrow$ & $\rho \!\uparrow$ & $cpu \!\downarrow$ \\ \midrule

\multirow{2}{*}{KIE} & \hspace{2.16pt}63.3 & \hspace{2.16pt}6.36 & -- & -- & \hspace{2.16pt}63.5 & \hspace{2.16pt}6.35 & -- & -- & \hspace{2.16pt}64.5 & \hspace{2.16pt}6.34 & -- & -- \\
 & (8.96) & (0.40) & -- & -- & (8.92) & (0.45) & -- & -- & (10.9) & (0.52) & -- & -- \\
\multirow{2}{*}{FIE} & \hspace{2.16pt}\bf{56.47} & \hspace{2.16pt}\bf{5.38} & \hspace{2.16pt}0.80 & \hspace{2.16pt}1.54  & \hspace{2.16pt}\bf{59.8} & \hspace{2.16pt}\bf{5.53} & \hspace{2.16pt}0.71 & \hspace{2.16pt}1.45 & \hspace{2.16pt}\bf{62.3} & \hspace{2.16pt}\bf{5.64} & \hspace{2.16pt}0.64 & 1.50 \\
 & (12.2) & (0.60) & -- & (0.34) & (13.4) & (0.62) & -- & (0.33) & (13.5) & (0.65) & -- & (0.34) \\
\multirow{2}{*}{K$^2$IE} & \hspace{2.16pt}\bf{53.0} & \hspace{2.16pt}\bf{5.54} & \hspace{2.16pt}0.97 & \hspace{2.16pt}0.16  & \hspace{2.16pt}\bf{55.1} & \hspace{2.16pt}\bf{5.63} & \hspace{2.16pt}0.90 & \hspace{2.16pt}0.14 & \hspace{2.16pt}\bf{57.9} & \hspace{2.16pt}\bf{5.77} & \hspace{2.16pt}0.85 & 0.13 \\
 & (10.2) & (0.49) & -- & (0.03)  & (11.1) & (0.51) & -- & (0.02) & (12.2) & (0.55) & -- & (0.03) \\

\toprule
\end{tabular}
} 
\end{center}
\end{table*}

In accordance with previous studies \citep{adams09, john18, aglietti19, kim2021}, we generated 1D datasets based on three types of intensity functions:
\begin{align}
\lambda^1_{\text{1D}}(x) &= 2 e^{-x/15} + e^{-[(x -25)/10]^2}, \ \ & \mathcal{X} &= [0,50], \notag \\
\lambda^2_{\text{1D}}(x) &= 5\sin(x^2)+6, & \mathcal{X} &= [0,5], \notag \\
\lambda^3_{\text{1D}}(x) &= \text{piecewise linear function}, & \mathcal{X} &= [0,100],
\end{align}
where $\lambda^3_{\text{1D}}(x)$ passes through the points: $(0, 2)$, $(25, 3)$, $(50, 1)$, $(75, 2.5)$, and $(100, 3)$. Furthermore, to evaluate the scalability of K$^2$IE and FIE with respect to data size, we generated 1D datasets using intensity functions scaled by a factor of ten, denoted by $10\!\times\!\lambda^q_{\text{1D}}(x)$ for $q \in \{1,2,3 \}$. For each intensity function, we simulated $100$ trial sequences and performed intensity estimation $100$ times using the compared methods. 

Table \ref{table_1d_synthetic} displays the predictive performance on the 1D synthetic datasets. It shows that our proposal K$^2$IE matched the predictive performances of FIE across all three datasets, while achieving significantly faster model fitting in terms of CPU time. K$^2$IE achieved comparable predictive performances with KIE on $\lambda^2_{\text{1D}}(x)$, but was outperformed by KIE on $\lambda^1_{\text{1D}}(x)$ and $\lambda^3_{\text{1D}}(x)$. This result is consistent with \cite{flaxman17}, demonstrating that KIE performed very well in low-dimensional settings. It is worth noting that the discrepancy of predictive performances between KIE and K$^2$IE became negligible on $10\!\times\!\lambda^{1,2,3}_{\text{1D}}(x)$, where the dataset size increases. Also, Table \ref{table_1d_synthetic} demonstrates that the CPU time of FIE increases with the data size, while that of K$^2$IE remains nearly constant. Figure \ref{fig_1D} displays some estimation results. 

\subsection{2D Synthetic Data}

Following the procedure in \cite{lloyd15}, we generated a 2D dataset from a sigmoidal Gaussian Cox process. Specifically, we first sampled a 2D function from a Gaussian process with an RKHS kernel, $k(\bm{x}, \bm{x}') = e^{-|\bm{x}-\bm{x}'|^2/2}$, over the domain $\mathcal{X} = [0,5]\times [0,5]$. The intensity function was then obtained by applying a sigmoid link function, $\sigma(z) = 50/(1+e^{-20z})$, to the sampled function. Using the intensity function, we simulated $100$ trials of event data and conducted intensity estimation $100$ times using the compared methods. The resulting dataset contained approximately 540 data points per trial.

In this study, we considered a scenario where the observation domain was divided into $5\!\times\!5\!=\!25$ sub-domains, 
\begin{equation}
\mathcal{X} = \bigcup_{j=1}^{25} \mathcal{X}_j, \ \ \mathcal{X}_j : \text{evenly partitioned 2D domain},
\end{equation}
with some of the sub-domains being missing. For each trial of the dataset, we randomly selected each sub-domain with a probability $p \in \{ 1.0, 0.9, 0.8 \}$, thereby generating three datasets, denoted as $\lambda^{1.0}_{\text{2D}}(x)$, $\lambda^{0.9}_{\text{2D}}(x)$, and $\lambda^{0.8}_{\text{2D}}(x)$, respectively.

Table \ref{table_2d_synthetic} displays the predictive performance on the 2D synthetic datasets, which shows that K$^2$IE and FIE consistently outperformed KIE in all datasets. This result suggests that K$^2$IE and FIE could more effectively handle edge effects than KIE in multi-dimensional settings. Notably, regarding the integrated squared error $L^2$, K$^2$IE achieved superior predictive performance, on average, than FIE, despite both methods employing the same equivalent kernels (with hyperparameters optimized individually for each model). It might be due to the fact that KIE is based on the minimization of the least squares loss (see Section \ref{sec_k2ie}). Another possible explanation for this result is that the optimization of the dual coefficient required in FIE may become unstable. Indeed, \citet{john18} reported that FIE can yield unreasonable solutions for highly modulating intensity functions. In contrast, K$^2$IE is expected to work more robustly, as it does not require the optimization of dual coefficients. Figure \ref{fig_1D} displays some estimation results on $\lambda_{\text{2D}}^{1.0}(x)$ and $\lambda_{\text{2D}}^{0.8}(x)$.

Additional experiments with a scalable Bayesian approach and on a real-world dataset are provided in Appendix \ref{app_add}.

\section{Discussions}\label{sec_conclusion}

We have proposed a novel penalized least squares loss formulation for estimating intensity functions that resides in an RKHS. Through the path integral representation of the squared Hilbert space norm, we showed that the optimization problem encompasses a representer theorem, and derived a feasible intensity estimator based on kernel methods. We evaluated the proposed estimator on synthetic data, confirming that it achieved comparable predictive accuracy while being substantially faster than the state-of-the-art kernel method-based estimator.

\subsubsection*{Limitations and Future Work}

As noted at the end of Section~\ref{sec_k2ie}, a key limitation of our K$^2$IE lies in its lack of a general guarantee for the non-negativity of the resulting intensity function. To investigate the effect, we conducted an analysis of how frequently K$^2$IE produces negative values using the 2D synthetic dataset $\lambda_{\text{2D}}^{1.0}$. Specifically, we evaluated the estimated intensity values at 500 $\times$ 500 grid points within the observation domain and computed the ratio of negative values. The mean $\pm$ standard deviation of this ratio across 100 trials was 0.059 $\pm$ 0.016, indicating that K$^2$IE can indeed produce negative estimates in practice--particularly in regions with sparse data--highlighting the necessity of a post-hoc correction such as clipping via $\max(\hat{\lambda}(\bm{x}), 0)$ in applications where negative intensity values are not permitted. As a direction for future work, we explore the technical challenges involved in incorporating non-negativity constraints directly into the functional optimization problem defined in Equation~(\ref{eq_lsf}).

One natural approach is to model the intensity function as a non-negative transformation $\sigma(f(\bm{x}))$ of a latent function $f(\bm{x})$ residing in an RKHS. In this setting, the functional analysis of the objective in Equation~(\ref{eq_lsf}) yields the following condition that the optimal function $\hat{f}(\bm{x})$ must satisfy:
\begin{equation*}
\begin{split}
\frac{1}{\gamma} f(\bm{x}) + \int_{\mathcal{X}} k(\bm{x},\bm{s})\sigma(f(\bm{s})) &\sigma'(f(\bm{s})) d\bm{s} \\
&= \sum_n k(\bm{x},\bm{x}_n) \sigma'(f(\bm{x}_n)),
\end{split}
\end{equation*}
where $\sigma'(y)$ $=$ $\frac{d\sigma}{dy}(y)$. When $\sigma(y)$ $=$ $y$, the above equation reduces to a Fredholm integral equation, for which Theorem \ref{thm_k2ie} provides a tractable solution. However, if $\sigma$ is nonlinear, even for simple cases like $\sigma(y) = y^2$, the resulting nonlinear integral equation becomes analytically and numerically challenging to solve. 

An alternative approach is to impose non-negativity constraints at a finite set of virtual points, which leads to a dual optimization problem. Although this approach may reduce the risk of negative estimates at the virtual points, it neither guarantees global non-negativity nor preserves the computational simplicity of K$^2$IE, due to the added complexity introduced by the dual optimization.

Does K$^2$IE truly fail to guarantee non-negativity in its original form? Interestingly, a sufficient condition to ensure the non-negativity of the equivalent kernels arising in Flaxman's estimator (and, of course, in K$^2$IE) has been established by \citet{kim2024inverse}. Specifically, when RKHS kernels belong to the class of inverse M-kernels (IMKs), the associated equivalent kernels $h(\bm{x},\bm{x}')$ are guaranteed to be non-negative. This suggests that K$^2$IE, as defined by a sum of equivalent kernels in Equation~(\ref{eq_k2ie}), may be a non-negative intensity estimator whenever the RKHS kernel is an IMK. In one-dimensional cases, the Laplace kernel is known to be an IMK, but no general construction of IMKs is currently known in higher dimensions--posing an intriguing open problem.

\section*{Impact Statement}

This paper presents work whose goal is to advance the field of Machine Learning. There are many potential societal consequences of our work, none of which we feel must be specifically highlighted here.

\bibliography{reference}

\begin{thebibliography}{51}
\providecommand{\natexlab}[1]{#1}
\providecommand{\url}[1]{\texttt{#1}}
\expandafter\ifx\csname urlstyle\endcsname\relax
  \providecommand{\doi}[1]{doi: #1}\else
  \providecommand{\doi}{doi: \begingroup \urlstyle{rm}\Url}\fi

\bibitem[Adams et~al.(2009)Adams, Murray, and MacKay]{adams09}
Adams, R.~P., Murray, I., and MacKay, D.~J.
\newblock Tractable nonparametric {B}ayesian inference in {P}oisson processes
  with {G}aussian process intensities.
\newblock In {\it International Conference on Machine Learning}, pp.\  9--16,
  2009.

\bibitem[Aglietti et~al.(2019)Aglietti, Bonilla, Damoulas, and
  Cripps]{aglietti19}
Aglietti, V., Bonilla, E.~V., Damoulas, T., and Cripps, S.
\newblock Structured variational inference in continuous {C}ox process models.
\newblock In {\it Advances in Neural Information Processing Systems 32}, 2019.

\bibitem[Atkinson(2010)]{atkinson2010personal}
Atkinson, K.
\newblock A personal perspective on the history of the numerical analysis of
  {F}redholm integral equations of the second kind.
\newblock In {\it The Birth of Numerical Analysis}, pp.\  53--72. World
  Scientific, 2010.

\bibitem[Bacry et~al.(2020)Bacry, Bompaire, Ga{\"\i}ffas, and
  Muzy]{bacry2020sparse}
Bacry, E., Bompaire, M., Ga{\"\i}ffas, S., and Muzy, J.-F.
\newblock Sparse and low-rank multivariate {H}awkes processes.
\newblock {\it Journal of Machine Learning Research}, 21\penalty0
  (50):\penalty0 1--32, 2020.

\bibitem[Cai et~al.(2024)Cai, Zhang, and Guan]{cai2024latent}
Cai, B., Zhang, J., and Guan, Y.
\newblock Latent network structure learning from high-dimensional multivariate
  point processes.
\newblock {\it Journal of the American Statistical Association}, 119\penalty0
  (545):\penalty0 95--108, 2024.

\bibitem[Clark et~al.(2003)Clark, Bradburn, Love, and
  Altman]{clark2003survival}
Clark, T.~G., Bradburn, M.~J., Love, S.~B., and Altman, D.~G.
\newblock Survival analysis part i: basic concepts and first analyses.
\newblock {\it British Journal of Cancer}, 89\penalty0 (2):\penalty0 232--238,
  2003.

\bibitem[Cox(1972)]{cox1972regression}
Cox, D.~R.
\newblock Regression models and life-tables.
\newblock {\it Journal of the Royal Statistical Society: Series B
  (Methodological)}, 34\penalty0 (2):\penalty0 187--202, 1972.

\bibitem[Cronie et~al.(2024)Cronie, Moradi, and Biscio]{cronie2024cross}
Cronie, O., Moradi, M., and Biscio, C.~A.
\newblock A cross-validation-based statistical theory for point processes.
\newblock {\it Biometrika}, 111\penalty0 (2):\penalty0 625--641, 2024.

\bibitem[Cunningham et~al.(2007)Cunningham, Byron, Shenoy, and
  Sahani]{cunningham07}
Cunningham, J.~P., Byron, M.~Y., Shenoy, K.~V., and Sahani, M.
\newblock Inferring neural firing rates from spike trains using {G}aussian
  processes.
\newblock In {\it Advances in Neural Information Processing Systems 20}, 2007.

\bibitem[Daley \& Vere-Jones(1988)Daley and Vere-Jones]{daley2006introduction}
Daley, D.~J. and Vere-Jones, D.
\newblock {\it An Introduction to the Theory of Point Processes}.
\newblock Springer-Verlag, New York, 1988.

\bibitem[Davis et~al.(2011)Davis, Lii, and Politis]{davis2011remarks}
Davis, R.~A., Lii, K.-S., and Politis, D.~N.
\newblock Remarks on some nonparametric estimates of a density function.
\newblock {\it Selected Works of Murray Rosenblatt}, pp.\  95--100, 2011.

\bibitem[Diggle(1985)]{diggle1985kernel}
Diggle, P.
\newblock A kernel method for smoothing point process data.
\newblock {\it Journal of the Royal Statistical Society: Series C (Applied
  Statistics)}, 34\penalty0 (2):\penalty0 138--147, 1985.

\bibitem[Diggle et~al.(2013)Diggle, Moraga, Rowlingson, and Taylor]{diggle13}
Diggle, P.~J., Moraga, P., Rowlingson, B., and Taylor, B.~M.
\newblock Spatial and spatio-temporal log-{G}aussian {C}ox processes: extending
  the geostatistical paradigm.
\newblock {\it Statistical Science}, 28\penalty0 (4):\penalty0 542--563, 2013.

\bibitem[Donner \& Opper(2018)Donner and Opper]{donner18}
Donner, C. and Opper, M.
\newblock Efficient {B}ayesian inference of sigmoidal {G}aussian {C}ox
  processes.
\newblock {\it Journal of Machine Learning Research}, 19:\penalty0 1--34, 2018.

\bibitem[Flaxman et~al.(2017)Flaxman, Teh, and Sejdinovic]{flaxman17}
Flaxman, S., Teh, Y.~W., and Sejdinovic, D.
\newblock Poisson intensity estimation with reproducing kernels.
\newblock In {\it Artificial Intelligence and Statistics}, pp.\  270--279.
  PMLR, 2017.

\bibitem[Gatrell et~al.(1996)Gatrell, Bailey, Diggle, and
  Rowlingson]{gatrell1996spatial}
Gatrell, A.~C., Bailey, T.~C., Diggle, P.~J., and Rowlingson, B.~S.
\newblock Spatial point pattern analysis and its application in geographical
  epidemiology.
\newblock {\it Transactions of the Institute of British Geographers}, pp.\
  256--274, 1996.

\bibitem[Gunter et~al.(2014)Gunter, Lloyd, Osborne, and Roberts]{gunter14}
Gunter, T., Lloyd, C., Osborne, M.~A., and Roberts, S.~J.
\newblock Efficient {B}ayesian nonparametric modelling of structured point
  processes.
\newblock In {\it Uncertainty in Artificial Intelligence}, 2014.

\bibitem[Hansen et~al.(2015)Hansen, Reynaud-Bouret, and
  Rivoirard]{hansen2015lasso}
Hansen, N.~R., Reynaud-Bouret, P., and Rivoirard, V.
\newblock Lasso and probabilistic inequalities for multivariate point
  processes.
\newblock {\it Bernoulli}, 21\penalty0 (1):\penalty0 83--143, 2015.

\bibitem[Heikkinen \& Arjas(1999)Heikkinen and Arjas]{heikkinen1999modeling}
Heikkinen, J. and Arjas, E.
\newblock Modeling a {P}oisson forest in variable elevations: A nonparametric
  {B}ayesian approach.
\newblock {\it Biometrics}, 55\penalty0 (3):\penalty0 738--745, 1999.

\bibitem[Holm(1979)]{holm1979simple}
Holm, S.
\newblock A simple sequentially rejective multiple test procedure.
\newblock {\it Scandinavian Journal of Statistics}, pp.\  65--70, 1979.

\bibitem[Hubbel \& Foster(1983)Hubbel and Foster]{sp1983diversity}
Hubbel, S.~P. and Foster, R.~B.
\newblock Diversity of canopy trees in a neotropical forest and implications
  for conservation.
\newblock {\it Tropical Rain Forest: Ecology and Management}, pp.\  25--41,
  1983.

\bibitem[John \& Hensman(2018)John and Hensman]{john18}
John, S.~T. and Hensman, J.
\newblock Large-scale {C}ox process inference using variational {F}ourier
  features.
\newblock In {\it International Conference on Machine Learning}, volume~80,
  pp.\  2362--2370. PMLR, 2018.

\bibitem[Jones(1993)]{jones1993simple}
Jones, M.~C.
\newblock Simple boundary correction for kernel density estimation.
\newblock {\it Statistics and Computing}, 3:\penalty0 135--146, 1993.

\bibitem[Kim(2021)]{kim2021}
Kim, H.
\newblock Fast {B}ayesian inference for {G}aussian {C}ox processes via path
  integral formulation.
\newblock In {\it Advances in Neural Information Processing Systems 34}, 2021.

\bibitem[Kim(2024)]{kim2024inverse}
Kim, H.
\newblock Inverse {M}-kernels for linear universal approximators of
  non-negative functions.
\newblock In {\it Advances in Neural Information Processing Systems 37}, 2024.

\bibitem[Kim et~al.(2022)Kim, Asami, and Toda]{kim2022fastbayesian}
Kim, H., Asami, T., and Toda, H.
\newblock Fast {B}ayesian estimation of point process intensity as function of
  covariates.
\newblock In {\it Advances in Neural Information Processing Systems 35}, 2022.

\bibitem[Kingma \& Ba(2014)Kingma and Ba]{kingma2014}
Kingma, D.~P. and Ba, J.
\newblock Adam: A method for stochastic optimization.
\newblock {\it arXiv preprint arXiv:1412.6980}, 2014.

\bibitem[Kowalczuk \& Kozlowski(1998)Kowalczuk and
  Kozlowski]{kowalczuk1998integrated}
Kowalczuk, Z. and Kozlowski, J.
\newblock Integrated squared error and integrated absolute error in recursive
  identification of continuous-time plants.
\newblock In {\it UKACC International Conference on Control'98 (Conf. Publ. No.
  455)}, volume~1, pp.\  693--698. IET, 1998.

\bibitem[Lai \& Xie(2006)Lai and Xie]{lai2006stochastic}
Lai, C.~D. and Xie, M.
\newblock {\it Stochastic Ageing and Dependence for Reliability}.
\newblock Springer Science \& Business Media, 2006.

\bibitem[L{\'a}nczky \& Gy{\H{o}}rffy(2021)L{\'a}nczky and
  Gy{\H{o}}rffy]{lanczky2021web}
L{\'a}nczky, A. and Gy{\H{o}}rffy, B.
\newblock Web-based survival analysis tool tailored for medical research
  (kmplot): development and implementation.
\newblock {\it Journal of Medical Internet Research}, 23\penalty0 (7):\penalty0
  e27633, 2021.

\bibitem[Lloyd et~al.(2015)Lloyd, Gunter, Osborne, and Roberts]{lloyd15}
Lloyd, C., Gunter, T., Osborne, M., and Roberts, S.
\newblock Variational inference for {G}aussian process modulated {P}oisson
  processes.
\newblock In {\it International Conference on Machine Learning}, volume~37,
  pp.\  1814--1822. PMLR, 2015.

\bibitem[Mercer(1909)]{mercer1909xvi}
Mercer, J.
\newblock Xvi. functions of positive and negative type, and their connection
  the theory of integral equations.
\newblock {\it Philosophical Transactions of the Royal Society of London.
  Series A, containing papers of a mathematical or physical character},
  209\penalty0 (441-458):\penalty0 415--446, 1909.

\bibitem[M{\o}ller et~al.(1998)M{\o}ller, Syversveen, and
  Waagepetersen]{moller98}
M{\o}ller, J., Syversveen, A.~R., and Waagepetersen, R.~P.
\newblock Log {G}aussian {C}ox processes.
\newblock {\it Scandinavian Journal of Statistics}, 25\penalty0 (3):\penalty0
  451--482, 1998.

\bibitem[Ogata(1988)]{ogata1988statistical}
Ogata, Y.
\newblock Statistical models for earthquake occurrences and residual analysis
  for point processes.
\newblock {\it Journal of the American Statistical Association}, 83\penalty0
  (401):\penalty0 9--27, 1988.

\bibitem[Parzen(1962)]{parzen1962estimation}
Parzen, E.
\newblock On estimation of a probability density function and mode.
\newblock {\it The Annals of Mathematical Statistics}, 33\penalty0
  (3):\penalty0 1065--1076, 1962.

\bibitem[Polyanin \& Manzhirov(1998)Polyanin and Manzhirov]{polyanin98}
Polyanin, A.~D. and Manzhirov, A.~V.
\newblock {\it Handbook of Integral Equations}.
\newblock CRC press, 1998.

\bibitem[Rahimi \& Recht(2007)Rahimi and Recht]{rahimi2007random}
Rahimi, A. and Recht, B.
\newblock Random features for large-scale kernel machines.
\newblock In {\it Advances in Neural Information Processing Systems 20}, 2007.

\bibitem[Ramlau-Hansen(1983)]{ramlau1983smoothing}
Ramlau-Hansen, H.
\newblock Smoothing counting process intensities by means of kernel functions.
\newblock {\it The Annals of Statistics}, pp.\  453--466, 1983.

\bibitem[Rathbun \& Cressie(1994)Rathbun and Cressie]{rathbun94}
Rathbun, S.~L. and Cressie, N.
\newblock Asymptotic properties of estimators for the parameters of spatial
  inhomogeneous {P}oisson point processes.
\newblock {\it Advances in Applied Probability}, 26\penalty0 (1):\penalty0
  122--154, 1994.

\bibitem[Scholkopf \& Smola(2018)Scholkopf and Smola]{scholkopf2018learning}
Scholkopf, B. and Smola, A.~J.
\newblock {\it Learning with Kernels: Support Vector Machines, Regularization,
  Optimization, and Beyond}.
\newblock MIT press, 2018.

\bibitem[Sch{\"o}lkopf et~al.(2001)Sch{\"o}lkopf, Herbrich, and
  Smola]{scholkopf2001generalized}
Sch{\"o}lkopf, B., Herbrich, R., and Smola, A.~J.
\newblock A generalized representer theorem.
\newblock In {\it International Conference on Computational Learning Theory},
  pp.\  416--426. Springer, 2001.

\bibitem[Sellier \& Dellaportas(2023)Sellier and
  Dellaportas]{sellier2023sparse}
Sellier, J. and Dellaportas, P.
\newblock Sparse spectral {B}ayesian permanental process with generalized
  kernel.
\newblock In {\it International Conference on Artificial Intelligence and
  Statistics}, pp.\  2769--2791. PMLR, 2023.

\bibitem[Shawe-Taylor \& Cristianini(2004)Shawe-Taylor and
  Cristianini]{shawe2004kernel}
Shawe-Taylor, J. and Cristianini, N.
\newblock {\it Kernel Methods for Pattern Analysis}.
\newblock Cambridge University Press, 2004.

\bibitem[Silverman(2018)]{silverman2018density}
Silverman, B.~W.
\newblock {\it Density Estimation for Statistics and Data Analysis}.
\newblock Routledge, 2018.

\bibitem[Teng et~al.(2017)Teng, Nathoo, and Johnson]{teng2017bayesian}
Teng, M., Nathoo, F., and Johnson, T.~D.
\newblock Bayesian computation for log-{G}aussian {C}ox processes: {A}
  comparative analysis of methods.
\newblock {\it Journal of Statistical Computation and Simulation}, 87:\penalty0
  2227--2252, 2017.

\bibitem[Tsuchida et~al.(2024)Tsuchida, Ong, and Sejdinovic]{tsuchida2024exact}
Tsuchida, R., Ong, C.~S., and Sejdinovic, D.
\newblock Exact, fast and expressive {P}oisson point processes via squared
  neural families.
\newblock In {\it Proceedings of the AAAI Conference on Artificial
  Intelligence}, volume~38, pp.\  20559--20566, 2024.

\bibitem[van~de Geer(2000)]{geer2000empirical}
van~de Geer, S.
\newblock {\it Empirical Processes in M-estimation}, volume~6.
\newblock Cambridge University Press, 2000.

\bibitem[Wahba(1990)]{wahba90}
Wahba, G.
\newblock {\it Spline Models for Observational Data}, volume~59.
\newblock SIAM, 1990.

\bibitem[Walder \& Bishop(2017)Walder and Bishop]{walder17}
Walder, C.~J. and Bishop, A.~N.
\newblock Fast {B}ayesian intensity estimation for the permanental process.
\newblock In {\it International Conference on Machine Learning}, volume~70,
  pp.\  3579--3588. PMLR, 2017.

\bibitem[Williams \& Seeger(2000)Williams and Seeger]{williams2000using}
Williams, C. and Seeger, M.
\newblock Using the {N}ystr{\"o}m method to speed up kernel machines.
\newblock In {\it Advances in Neural Information Processing Systems 13}, 2000.

\bibitem[Yang et~al.(2014)Yang, Sindhwani, Avron, and Mahoney]{yang2014quasi}
Yang, J., Sindhwani, V., Avron, H., and Mahoney, M.
\newblock Quasi-{M}onte {C}arlo feature maps for shift-invariant kernels.
\newblock In {\it International Conference on Machine Learning}, pp.\
  485--493. PMLR, 2014.

\end{thebibliography}
\bibliographystyle{icml2025_2}


\appendix

\section{Explanation of the Least Squares Loss}\label{app_ls}

Let $\mathbb{E}$ denote the expectation with respect to data points generated from the true intensity function $\lambda^*(\bm{x})$. We consider the expected integrated squared loss between the estimator $\hat{\lambda}(\bm{x})$ and the true intensity function $\lambda^*(\bm{x})$, defined as:
\begin{equation*}
\begin{split}
\mathbb{E} \Bigl[ \int_{\mathcal{X}} \bigl|\hat{\lambda}&(\bm{x}) - \lambda^*(\bm{x}) \bigr|^2 d\bm{x} \Bigr] = \mathbb{E} \Bigl[ \int_{\mathcal{X}} \hat{\lambda}^2(\bm{x}) d\bm{x} \Bigr] \\
& - 2\mathbb{E} \Bigl[ \int_{\mathcal{X}} \hat{\lambda}(\bm{x}) \lambda^*(\bm{x}) d\bm{x} \Bigr] + \mathbb{E} \Bigl[ \int_{\mathcal{X}} \lambda^{*2}(\bm{x}) d\bm{x} \Bigr].
\end{split}
\end{equation*}
The third term on the right-hand side is independent of the estimator and can, therefore, be omitted. The second term can be decomposed as follows:
\begin{equation*}
\begin{split}
2\mathbb{E} \Bigl[ \int_{\mathcal{X}} &\hat{\lambda}(\bm{x}) \lambda^*(\bm{x}) d\bm{x} \Bigr] = 2\mathbb{E} \Bigl[ \int_{\mathcal{X}} \hat{\lambda}(\bm{x}) \sum_{n=1}^N \delta(\bm{x}-\bm{x}_n) d\bm{x} \Bigr] \\
&+ 2\mathbb{E} \Bigl[  \int_{\mathcal{X}} \hat{\lambda}(\bm{x}) \Bigl( \lambda^*(\bm{x}) - \sum_{n=1}^N \delta(\bm{x}-\bm{x}_n) \Bigr) d\bm{x} \Bigr],
\end{split}
\end{equation*}
where the second term on the right-hand side vanishes due to Campbell's theorem \cite{daley2006introduction}:
\begin{equation*}
\begin{split}
&\int_{\mathcal{X}} \mathbb{E} \Bigl[\hat{\lambda}(\bm{x})\Bigr] \lambda^*(\bm{x}) d\bm{x} - \sum_{n=1}^N \mathbb{E} \Bigl[\hat{\lambda}(\bm{x}_n)\Bigr] \\
&\ \ = \int_{\mathcal{X}} \mathbb{E} \Bigl[\hat{\lambda}(\bm{x})\Bigr] \lambda^*(\bm{x}) d\bm{x} - \int_{\mathcal{X}} \mathbb{E} \Bigl[\hat{\lambda}(\bm{x}) \Bigr] \lambda^*(\bm{x}) d\bm{x} = 0.
\end{split}
\end{equation*}
Combining the above expressions yields the following identity:
\begin{equation*}
\begin{split}
\mathbb{E} \Bigl[ \int_{\mathcal{X}} \bigl|\hat{\lambda}(\bm{x}) &- \lambda^*(\bm{x}) \bigr|^2 d\bm{x} \Bigr] \\
&= \mathbb{E} \Bigl[ \int_{\mathcal{X}} \hat{\lambda}^2(\bm{x}) d\bm{x} - 2\sum_{n=1}^N \hat{\lambda}(\bm{x}_n) \Bigr] + C,
\end{split}
\end{equation*}
where \( C \) is a constant. This shows that the least squares loss defined in (\ref{eq_ls}) corresponds to the empirical integrated squared loss.

\section{Proof of Theorem \ref{thm_k2ie} via Mercer's Theorem}\label{app_mercer}

We present a proof of Theorem~\ref{thm_k2ie} based on Mercer's Theorem, following an approach similar to that of \citet{flaxman17}.

\begin{proof}

Using the Mercer expansion of the RKHS kernel given in Equation~(\ref{eq_equiv0}), any function $\lambda \in \mathcal{H}_k$ can be expressed as $\lambda(\cdot) = \sum_{m} b_m e_m(\cdot)$, where $\{b_m\}_m$ are the expansion coefficients and the RKHS norm is given by $||\lambda||^2_{\mathcal{H}_k} = \sum_m b_m^2/\eta_m < \infty$. Substituting this into the objective in Equation~(\ref{eq_lsf}), we obtain:
\begin{equation*}
\begin{split}
- 2&\sum_{n=1}^N \lambda(\bm{x}_n) + \frac{1}{\gamma} ||\lambda||^2_{\mathcal{H}_k} + \int_{\mathcal{X}} \lambda(\bm{x})^2 d\bm{x} \\
&= - 2\sum_{n=1}^N \lambda(\bm{x}_n) + \frac{1}{\gamma} \sum_m b_m^2/\eta_m \\
&\quad + \sum_m \sum_{m'} b_m b_{m'} \int_{\mathcal{X}} e_m(\bm{x}) e_{m'}(\bm{x}) d\bm{x} \\
&= - 2\sum_{n=1}^N \lambda(\bm{x}_n) + \frac{1}{\gamma} \sum_m b_m^2/\eta_m + \sum_m b_m^2 \\
&= - 2\sum_{n=1}^N \lambda(\bm{x}_n) + \sum_m \Bigl( \frac{\eta_m}{\eta_m + 1/\gamma} \Bigr)^{-1} b_m^2,
\end{split} 
\end{equation*}
where the orthogonality condition, $\int_{\mathcal{X}}e_m(\bm{x})e_{m'}(\bm{x})d\bm{x}$ $=$ $\delta_{mm'}$, is used.
The above equation shows that if we define a new RKHS kernel $q(\cdot,\cdot)$ as
\begin{equation*}
q(\bm{x},\bm{x}') = \sum_{m=1}^{\infty} \frac{\eta_m}{\eta_m + 1/\gamma} e_m(\bm{x}) e_m(\bm{x}'),
\end{equation*}
the optimization problem in Equation~(\ref{eq_lsf}) reduces to:
\begin{equation*}
\min_{\lambda \in \mathcal{H}_q} \biggl\{ - 2\sum_{n=1}^N \lambda(\bm{x}_n) + ||\lambda||^2_{\mathcal{H}_q}\biggr\},
\end{equation*}
where $||\cdot||^2_{\mathcal{H}_q}$ represents the squared norm of an RKHS $\mathcal{H}_q$ associated with $q(\cdot,\cdot)$. By construction, $q(\cdot,\cdot)$ coincides with the equivalent kernel defined in Equation~(\ref{eq_equiv0}).
According to the classical representer theorem~\citep{scholkopf2001generalized}, the optimal solution to this problem lies in the span of kernel evaluations at the data points:
\begin{equation*}
\hat{\lambda}(\bm{x}) = \sum_{n=1}^N \alpha_n q(\bm{x},\bm{x}_n),
\end{equation*}
where the dual coefficients $\bm{\alpha} = (\alpha_1, \dots, \alpha_N)^\top$ minimize the objective. Taking the gradient of the objective with respect to $\bm{\alpha}$ yields:
\begin{equation*}
\begin{split}
&\frac{\partial}{\partial\alpha_n} \biggl[ - 2\sum_{n=1}^N \lambda(\bm{x}_n) + ||\lambda||^2_{\mathcal{H}_q} \biggr] \\
& = -2 \sum_{n'=1}^N q(\bm{x}_{n'},\bm{x}_n) + 2 \alpha_n \sum_{n'=1}^N q(\bm{x}_{n'},\bm{x}_n) = 0, \\
&\therefore \ \ \alpha_n = 1.
\end{split}
\end{equation*}
This completes the proof.
\end{proof}

\section{Extension of Theorem \ref{thm_k2ie}}\label{app_ext}
\renewcommand{\theequation}{C\arabic{equation}}
\setcounter{equation}{0}

\begin{proposition}[] \label{prop_k2ie}
Let $k : \mathbb{R}^d \times \mathbb{R}^d$ $\rightarrow$ $\mathbb{R}$ be a continuous positive semi-definite kernel. Then the solution $\hat{\lambda}(\cdot)$ to the optimization problem~(\ref{eq_lsf}) admits a representer theorem with respect to a transformed RKHS kernel $h(\cdot, \cdot)$, which is defined via the following Fredholm integral equation:
\begin{equation}\label{eq_app_h}
\begin{split}
\frac{1}{\gamma} h(\bm{x},\bm{x}') + \int_{\mathcal{X}} k(\bm{x},\bm{s}) h(\bm{s},\bm{x}') d\bm{s} = k(\bm{x},\bm{x}')&, \\
(\bm{x}, \bm{x}') \in \mathbb{R}^d \times \mathbb{R}^d&.
\end{split}
\end{equation}
Moreover, its dual coefficient is equal to unity:
\begin{equation*}
\hat{\lambda}(\bm{x}) = \sum_{n=1}^N h(\bm{x},\bm{x}_n), \qquad \bm{x} \in \mathbb{R}^d.
\end{equation*}
\end{proposition}

\begin{proof}
Let the integral operator associated with the RKHS kernel $k(\cdot,\cdot)$ be defined as $\mathcal{K}\cdot (\bm{x})$ $=$ $\int_{\mathbb{R}^d} \cdot~k(\bm{x},\bm{s}) d\bm{s}$, and its inverse operator be denoted by $\mathcal{K}^*  \cdot (\bm{x})$ $=$ $\int_{\mathbb{R}^d} \cdot~k^*(\bm{x},\bm{s}) d\bm{s}$. Using the path integral formulation of Gaussian processes \cite{kim2021}, the squared norm in the RKHS can be expressed in the functional form:
\begin{equation*}
 ||\lambda||^2_{\mathcal{H}_k}  = \iint_{\mathbb{R}^d\times\mathbb{R}^d} \!\!\!\!\!\! k^*(\bm{x},\bm{s}) \lambda(\bm{x}) \lambda(\bm{s}) d\bm{x} d\bm{s}. 
 \end{equation*}
Based on this representation, the objective functional in Equation~(\ref{eq_lsf}) becomes:
\begin{equation*}
S (\lambda) = - 2\sum_{n=1}^N \lambda(\bm{x}_n) + \iint_{\mathbb{R}^d\times\mathbb{R}^d} \!\!\!\!\!\! h^*(\bm{x},\bm{s}) \lambda(\bm{x}) \lambda(\bm{s}) d\bm{x} d\bm{s},
\end{equation*}
where $h^*(\cdot,\cdot)$ is defined in terms of $k^*(\cdot,\cdot)$, the Dirac delta function $\delta(\cdot)$, and the indicator function $\bm{1}_{(\cdot)}$,
\begin{equation}\label{eq_equiv_inv2}
h^*(\bm{x},\bm{s}) = \delta(\bm{x}-\bm{s})\bm{1}_{\bm{s} \in \mathcal{X}} + \frac{1}{\gamma}k^*(\bm{x},\bm{s}).
\end{equation}
The minimizer $\hat{\lambda}(\bm{x})$ of $S(\lambda)$ satisfies the equation obtained by setting the functional derivative to zero:
\begin{equation*}
\frac{\delta S}{\delta \hat{\lambda}} = -2 \sum_{n=1}^N \delta (\bm{x}-\bm{x}_n) + 2 \int_{\mathbb{R}^d} h^*(\bm{x},\bm{s}) \hat{\lambda}(\bm{s}) d\bm{s} = 0.
\end{equation*}
Define the integral operator corresponding to $h^*(\cdot, \cdot)$ by $\mathcal{H}^* \cdot (\bm{x})$ $=$ $\int_{\mathbb{R}^d} \cdot~h^*(\bm{x},\bm{s}) d\bm{s}$, and its inverse operator by $\mathcal{H} \cdot (\bm{x})$ $=$ $\int_{\mathbb{R}^d} \cdot~h(\bm{x},\bm{s}) d\bm{s}$.
Applying $\mathcal{H}$ to both sides of the functional equation yields:
\begin{equation*}
\hat{\lambda}(\bm{x}) = \sum_{n=1}^N h(\bm{x},\bm{x}_n), \qquad \bm{x} \in \mathbb{R}^d,
\end{equation*}
where we have used the identity, $(\mathcal{H} \mathcal{H}^*)\!\cdot\!(\bm{x})$ $=$ $\int_{\mathbb{R}^d} \cdot \delta(\bm{x}\!-\!\bm{s}) d\bm{s}$.
Furthermore, applying the operator $\mathcal{H}$ to Equation~(\ref{eq_equiv_inv2}) leads to the relation, 
\begin{equation*}
\delta(\bm{x}-\bm{x}') = h(\bm{x},\bm{x}')\bm{1}_{\bm{x} \in \mathcal{X}} + \frac{1}{\gamma}  \int_{\mathbb{R}^d} k^*(\bm{x},\bm{s}) h(\bm{s},\bm{x}') d\bm{s},
\end{equation*}
and subsequent application of the operator $\mathcal{K}$ results in
\begin{equation*}
\begin{split}
k(\bm{x},\bm{x}') = \int_{\mathcal{X}} k(\bm{x},\bm{s}) h(\bm{s},\bm{x}')d\bm{s} + \frac{1}{\gamma}  h(\bm{x},\bm{x}')&, \\
(\bm{x}, \bm{x}') \in \mathbb{R}^d \times \mathbb{R}^d&,
\end{split}
\end{equation*}
which is identical to Equation~(\ref{eq_app_h}).
\end{proof}

\section{Ablation Study on the Number of Random Features}\label{app_ablation}
\renewcommand{\thetable}{D\arabic{table}}
\setcounter{figure}{0}
\setcounter{table}{0}

\begin{table}[h]
\begin{center}
\caption{Predictive performance of K$^2$IE on the 2D synthetic data $\lambda_{\text{2D}}^{1.0}$ as a function of the number of feature maps. Brackets represent standard errors over 100 trials.}
\label{table_ab}
\fontsize{9.0pt}{10.0pt}\selectfont{
\begin{tabular}{ccccc}
\toprule
$2M$ & $20$ & $100$ & $300$ & $500$ \\ \midrule
\multirow{2}{*}{$\ L^2$} & 147 & 75.4 & 53.2 & 53.0 \\
& (8.28) & (10.4) & (10.7) & (10.2) \\ \midrule
\multirow{2}{*}{$|L|$} & 9.807 & 6.681 & 5.56 & 5.54 \\
& (0.26) & (0.45) & (0.52) & (0.49) \\
\toprule
\end{tabular}
}
\end{center}
\vskip -0.1in
\end{table}

We conducted an ablation study to investigate the effect of the number of random features ($2M$) on the predictive performance of K$^2$IE using the 2D synthetic dataset $\lambda_{\text{2D}}^{1.0}$. As shown in Table~\ref{table_ab}, both the integrated squared error and the integrated absolute error consistently decrease as $M$ increases. These results indicate that K$^2$IE benefits from more random features, and that the setting $2M$ $=$ $500$, used in Section~\ref{sec_experiment}, provides sufficiently accurate and stable estimates.

\section{Additional Experiments}\label{app_add}
\renewcommand{\thetable}{D\arabic{table}}
\setcounter{figure}{0}
\setcounter{table}{0}

\subsection{Comparison with a Variational Bayesian model}

We conducted an additional experiment on the 2D synthetic dataset $\lambda_{\text{2D}}^{1.0}$ to compare against a scalable Bayesian model. Here, we adopted a variational Bayesian approach based on a Gaussian Cox process with a quadratic link function \cite{lloyd15}, where a Gaussian RKHS kernel and 10 $\times$ 10 inducing points were employed. We employed a gradient descent algorithm, {\it Adam} \cite{kingma2014}, to perform the model fitting, where the number of iterations and the learning parameter were set as 5000 and 0.01, respectively. $L^2$, $|L|$, and $cpu$ achieved by the Bayesian model were 63.9 (12.2), 5.55 (0.46), and 51.8 (32.2), respectively, where standard deviations are in brackets. The result highlights the high efficiency of K$^2$IE.

\subsection{Comparison on a Real-world Dataset}
\renewcommand{\thetable}{E\arabic{table}}
\setcounter{figure}{0}
\setcounter{table}{0}

We conducted an additional experiment using an open 2D real-world dataset, {\it bei}, in the R package spatsta (GPL-3). It consists of locations of 3605 trees of the species {\it Beilschmiedia pendula} in a tropical rain forest \cite{sp1983diversity}. 

Following \cite{cronie2024cross}, we randomly labeled the data points with independent and identically distributed marks $\{1, 2, 3\}$ from a multinomial distribution with parameters $(p_1, p_2, p_3)$ $=$ $(0.3, 0.3, 0.7)$, and assigned the points with label 1 and 2 to training data and test data, respectively; we repeated it 100 times for evaluation. A $10\!\times\!10$ logarithmic grid search was conducted for $\gamma \in [0.001, 1]$ and $\bm{\beta} \in [0.1, 100] \cdot \overline{\bm{\beta}}$, where $  \overline{\bm{\beta}} = ( \overline{\beta}_1, \dots, \overline{\beta}_d)^{\top}$ for $\overline{\beta}_i = 1/\bigl[\max_j \bigl(X_{ij}^{\text{max}}\bigr)-\min_j \bigl(X_{ij}^{\text{min}}\bigr)\bigr]$. 

Let  the observation domain $\mathcal{X}$ be regularly devided into 10 $\times$ 10 sub-domains as $\mathcal{X}$ $=$ $\bigcup_{j=1}^{100} \mathcal{X}_j$. We evaluated the predictive performance of the estimator $\hat{\lambda}(x)$ based on the test least squares loss ($L_{s}$) and the test negative log-likelihood of counts ($L_{c}$): 
\begin{equation*}
\begin{split}
L_{s} &= \int_{\mathcal{X}} \hat{\lambda}^2(\bm{x}) d\bm{x} - 2\sum_{n \in D_{\text{test}}} \hat{\lambda}(\bm{x}_n), \\
L_{c} &= \sum_{j=1}^{100} \biggl[ \hat{\Lambda}_j \!- N_j \log \hat{\Lambda}_j \!+ \log \bigl( N_j ! \bigr) \biggr], \ \hat{\Lambda}_j  \!=\! \int_{\mathcal{X}_i} \!\!\hat{\lambda}(\bm{x}) d\bm{x},
\end{split}
\end{equation*}
where $D_{\text{test}}$ denotes the test data, and $N_j$ represents the number of  test data points observed within $\mathcal{X}_j$.
Table \ref{table_real} displays the results, showing that K$^2$IE achieved the best performance on $L_s$ but was outperformed by KIE on $L_c$, which could be because the hyperparameters were optimized based on the least squares loss and the log-likelihood for K$^2$IE and KIE, respectively. 

\begin{table}[]
\begin{center}
\caption{Results on the real-world data {\it bei} across 100 trials with standard errors in brackets. The notation follows Table \ref{table_1d_synthetic}.}
\label{table_real}
\fontsize{9.0pt}{10.0pt}\selectfont{
\begin{tabular}{ccccc}
\toprule
 & $L_{s}\!\downarrow$ & $L_c\!\downarrow$ & $cpu$ \\ \midrule
\multirow{2}{*}{KIE} & -5.80 & \bf{267} & -- \\
& (0.32) & (11.5) &--  \\ 
\multirow{2}{*}{FIE} & -5.16 & 287 & 5.15 \\
& (0.26) & (15.1) & (1.57)  \\
\multirow{2}{*}{K$^2$IE} & \bf{-6.16} & 279 & 0.17  \\
& (0.44) & (13.2) & (0.04)  \\
\toprule
\end{tabular}
}
\end{center}
\vskip -0.1in
\end{table}

\end{document}